%% file: iclr2027_conference.tex
\documentclass{article} % For LaTeX2e
\input{tex_engine_compat.tex}
\usepackage{iclr2027_conference,times}

\input{math_commands.tex}

\usepackage{hyperref}
\usepackage{url}

\usepackage{graphicx}
\usepackage{booktabs}
\usepackage{multirow}
\usepackage{amssymb}
\usepackage{caption}
\usepackage{xcolor}

\usepackage{graphicx}
\usepackage{booktabs}
\usepackage{multirow}
\ifdefined\TASAmarkchanges
  \newcommand{\rev}[1]{{\color{blue}#1}}
\else
  \newcommand{\rev}[1]{{\color{black}#1}}
\fi

\title{Think in Strategies, Act with Tools: Traceable Agentic Optimization for Combinatorial Problems}

\author{Zhongnian Li, Shi Xu, Peng Ying, Weikang Gong, Zhixiao Wang, Xinzheng Xu\thanks{Corresponding author: \texttt{xxzheng@cumt.edu.cn}}\\
School of Computer Science and Technology /
School of Artificial Intelligence\\
China University of Mining and Technology\\
Xuzhou, China\\
\texttt{\{zhongnianli,xushi,pengying,wkgong,zhxwang,xxzheng\}@cumt.edu.cn}}

\iclrfinalcopy

\begin{document}

\maketitle
\pagestyle{plain}

\begin{abstract}
Optimization by PROmpting (OPRO) directly leverages Large Language Models (LLMs) to generate candidate solutions through natural-language prompting. However, OPRO’s optimization process is difficult to track, which prevents LLMs from capturing useful information accumulated over previous iterations, thereby limiting solution quality. To address this limitation, we propose a Tool-Augmented Strategic Agent (TASA) which summarizes and exploits trajectory to improve solution quality. Specifically, TASA plans the strategy for each step based on the optimization trajectory, which decomposes the optimization process into traceable strategy selections.
Moreover, we leverage LLM to construct problem-specific tools and introduce a strategy-guided adaptive tool planner, which efficiently utilizes the information of trajectory to update solutions. We \rev{establish} local convergence under \rev{persistent exploration and evaluate the complete framework under a shared candidate-evaluation budget}. \rev{The} results \rev{support its} solution quality \rev{and the traceability of strategy-guided search}.
\end{abstract}

\section{Introduction}

Recent advances in Large Language Models (LLMs)~\cite{team2024gemini,grattafiori2024llama,guo2025deepseek,yang2025qwen3} substantially improve their capabilities in reasoning~\cite{wei2022cot,gao2023pal}, planning~\cite{yao2023tree,zhou2024language}, and decision-making~\cite{yao2023react}, providing possibilities for using LLMs to solve optimization problems. Optimization by PROmpting (OPRO)~\cite{yang2024OPRO} leverages an LLM as the optimizer to generate candidate solutions by iterative prompting process. Specifically, OPRO prompts the LLM to directly generate promising candidate solutions by constructing a prompt that contains a description of the optimization problem and previously evaluated solutions. Due to its advantages, OPRO attracts considerable attention, inspiring broader research on LLM-based optimization~\cite{guo2023towards,liu2024evolutionary,stein2025llamea,zheng2025mctsahd,ye2025llmlns}.

\begin{figure*}[t]
\vspace{-30pt}
\centering
\includegraphics[width=0.9\textwidth]{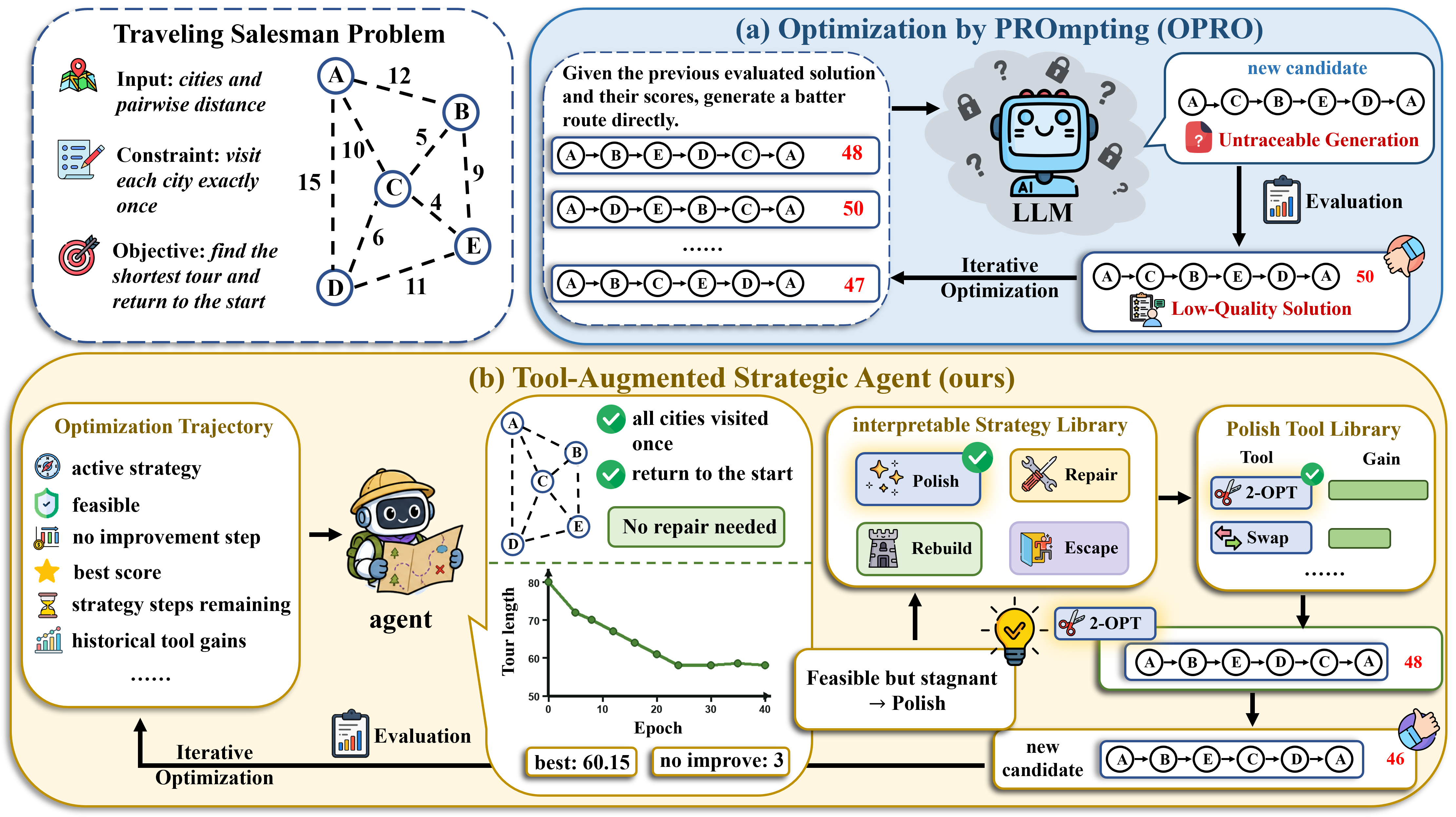} % Reduce the figure size so that it is slightly narrower than the column.
\caption{\emph{Motivation and overview of our Tool-Augmented Strategic Agent.} TASA shifts the role of the LLM from directly generating solutions to planning optimization strategies from optimization trajectories, and executes these strategies through strategy-guided adaptive tool planner.}
\vspace{-10pt}
\label{fig1}
\end{figure*}

Unfortunately, the limited traceability of OPRO’s optimization process makes it difficult for LLMs to capture useful information from previous iterations, thereby limiting solution quality.
As illustrated in Figure~\ref{fig1}(a), the LLM directly outputs a candidate solution without explicitly stating the intention or trajectory behind its generation. 
% Such information is crucial for LLMs to develop a coherent understanding of the optimization process and make informed decisions about subsequent exploration.
Consequently, OPRO retains only the candidate solutions generated across iterations, without an explicit account of how each new solution is generated. This lack of the information hinders LLM from accurately assessing the current search state, thereby limiting ability of OPRO to identify promising search directions. For example, when the candidate set is dominated by similar solutions, the OPRO may struggle to escape the corresponding region of the solution space and discover a better solution.

 To address this limitation, we propose a Tool-Augmented Strategic Agent (TASA) to plan the strategy for each step based on the optimization trajectory. Specifically, TASA constructs an explicit optimization trajectory by continuously recording information about its optimization process, primarily the selected optimization strategies and tool invocations. As illustrated in Figure~\ref{fig1}, after constructing the trajectory, the agent analyzes the recorded information and accordingly selects the polish strategy to further shorten the current tour through local adjustments. TASA aims to enable agent to exploit the optimization trajectory for more informed optimization decisions by shifting the LLM’s role from directly generating candidate solutions to planning optimization strategies.

In this paper, we leverage an LLM to construct a library of problem-specific tools according to the characteristics of each optimization problem. The resulting library organizes executable, problem-adaptive tools by optimization strategy. Building on this tool library, we introduce a strategy-guided adaptive tool planner, which efficiently translates the optimization strategies selected by agent into concrete solution updates. 
This mechanism balances the exploitation of previously well-performing tools, the exploration of underexplored tools, and the avoidance of repeatedly selecting poorly performing tools. Finally, the planner applies the selected tool to the candidate solution, enabling the efficient execution of the LLM-planned strategy.
Moreover, we analyze the probabilistic local convergence of TASA under strategy-conditioned tool neighborhoods. Experimental results across diverse combinatorial optimization problems demonstrate that TASA not only makes the optimization process traceable but also improves solution quality.

Our main contributions are summarized \rev{below}\rev{.}
\begin{itemize}
\item We propose a tool-augmented strategic agent for traceable combinatorial optimization and provide a theoretical analysis of its probabilistic local convergence under strategy-conditioned tool neighborhoods.
\item We construct problem-specific tools and introduce a strategy-guided adaptive tool planner that invokes these tools to efficiently update solutions.
\item Experimental results across diverse combinatorial optimization problems demonstrate that TASA not only makes the optimization process traceable but also improves solution quality.
\end{itemize}
\vspace{-10pt}

\section{Related Work}

\subsection{LLM-Based Combinatorial Optimization}

Large Language Models (LLMs)~\cite{grattafiori2024llama,team2024gemini,guo2025deepseek,yang2025qwen3} apply to combinatorial optimization through both direct solution generation and automatic heuristic design~\cite{daros2026survey}. Optimization by PROmpting (OPRO)~\cite{yang2024OPRO} treats the LLM as an optimizer and generates new candidates from previously evaluated solutions. FunSearch~\cite{romera2024FunSearch} combines a LLM with an evaluator to evolve program components in function space, while Evolution of Heuristics (EoH)~\cite{liu2024EoH} jointly evolves natural-language heuristic ideas and their executable implementations. These methods demonstrate the potential of LLMs to optimize over both candidate solutions and heuristic programs. \rev{TASA complements these approaches with explicit links between trajectory-conditioned strategies}, \rev{executable tool updates}, and \rev{observed search outcomes}.

\subsection{Tool-Augmented Agents}

Tool-augmented agents~\cite{karpas2022mrkl,paranjape2023art,
shen2023hugginggpt,song2023restgpt,hao2023toolkengpt,patil2024gorilla,qin2024toolllm,tang2023toolalpaca,lu2023chameleon,xu2023rewoo,wu2024toolplanner,kong2024tptu,gao2024confucius} extend LLMs by enabling them to interact with external tools and environments~\cite{cai2024toolmakers,qian2023creator, yuan2024craft}. ReAct~\cite{yao2023react} integrates reasoning with action generation, allowing LLMs to update their plans based on observations from the environment, while Toolformer~\cite{schick2023toolformer} trains language models to determine which APIs to invoke, when to invoke them, and how to incorporate their outputs. In optimization~\cite{li2023optiguide,ahmaditeshnizi2024optimus}, OptiGuide translates natural-language queries into code that modifies and invokes existing solvers, supporting solution interpretation and what-if analysis.

However, these methods mainly target general task interaction or solver-assisted analysis, rather than treating tools as structured operators for iterative solution improvement. In contrast, TASA constructs problem-specific tools and employs a strategy-guided adaptive tool planner that selects them according to optimization trajectory.

\section{Method}

\subsection{Problem Formalization}

In this paper, we consider a combinatorial optimization problem \(P\) defined over a finite discrete candidate space \(\Omega_P\). Since candidate solutions generated during optimization may violate problem constraints, we utilize \(v_P(x)\) to represent the constraint-violation status of a candidate \(x\). Each component of \(v_P(x)\) corresponds to a specific constraint, taking the value 0 if the constraint is satisfied and 1 if it is violated. The vector is divided into a hard-constraint component \(v_P^{\mathrm{hard}}(x)\) and a soft-constraint component \(v_P^{\mathrm{soft}}(x)\) (when the problem contain soft-constraint). Hard constraints determine the feasibility of a candidate, whereas soft constraints express desirable but nonmandatory requirements. A candidate is considered feasible if none of its hard constraints are violated. Accordingly, the feasible solution set is defined as
\begin{equation}
\small
\mathcal{X}_P
=
\left\{
x \in \Omega_P
\mid
v_P^{\mathrm{hard}}(x)=0
\right\}
\label{eq:feasible_set}
\end{equation}

To unify the optimization directions of different combinatorial optimization problems, we transform original objective function \(g_P(x)\) of each problem into a higher-is-better score by a problem-specific transformation function \(\phi_P(\cdot)\). The unified internal scoring function \(f_P(x)\) defined as
\begin{equation}
\small
f_P(x)
=
\phi_P\!\left(g_P(x)\right)
-
\mathcal{P}_P\!\left(v_P(x)\right)
\label{eq:unified_score}
\end{equation}
where \(\mathcal{P}_P(\cdot)\) is a problem-specific penalty function used to penalize constraint violations and guide the search toward the feasible region.

Under this formulation, our goal is to identify the best feasible solution found within a budget of \(M\) candidate evaluations \rev{as}
\begin{equation}
\small
\hat{x}_P^{M}
\in
\arg\max_{x\in\mathcal{X}_P\cap\mathcal{E}_M}
f_P(x)
\label{eq:budgeted_objective}
\end{equation}
where \(\mathcal{E}_M={x^{1},x^{2},\ldots,x^{M}}\) denotes the set of candidates evaluated during the search.

\subsection{Problem-Specific Tools Construction}
To maintain adaptability of TASA across tasks, we utilize an LLM to construct a problem-specific tools library based on the characteristics of each optimization problem. TASA constructs tool library through the following two coupled stages\rev{.}

\paragraph{Tool Archetype Generation.} Given a structured context containing the problem description, constraints, candidate-solution schema, and strategy definitions, the LLM generates tool archetypes for each of the four strategies. Each archetype describes the intended functionality and behavior rather than providing directly executable code.

\paragraph{Problem-Specific Tool Archetype Instantiation.} Once generated, the tool archetypes are validated for structural correctness, safety, and compatibility with the available primitive operators. Valid archetypes are then bound to registered operators and instantiated as callable tools adapted to the current problem, while invalid archetypes are corrected or rejected.

\rev{The LLM specifies each tool's strategy}, \rev{operator sequence}\rev{, and parameters}\rev{. Registered operators execute these} updates \rev{and may perform multiple updates per candidate}. \rev{Tool-construction prompts appear} in the supplementary material.

\subsection{Optimization Strategy Planning}

\rev{TASA uses an LLM to select optimization strategies from} the \rev{problem description}, \rev{available tools}\rev{, and optimization trajectory}\rev{. This allows} the \rev{agent to relate the} current search state \rev{to} the \rev{purpose of the next solution update}. \rev{We} let \(z\in\mathcal{Z}:= \{\texttt{rebuild},\texttt{repair},\texttt{polish},\texttt{escape}\} \) denote an optimization strategy\rev{, with the} four strategies \rev{defined as follows}\rev{.}
\vspace{-5pt}
\begin{itemize}
\item \textbf{rebuild} reconstructs a solution from a different structural configuration, enabling the exploration of a new region of the search space\rev{.}
\item \textbf{repair} focuses on reducing constraint violations and restoring the feasibility of the current solution\rev{.}
\item \textbf{polish} performs local refinement around a promising feasible solution to further improve score\rev{.}
\item \textbf{escape} applies stronger structural perturbations to move away from a stagnant search region.
\end{itemize}
\vspace{-5pt}

\rev{The allocation among} these intentions \rev{adapts} to the \rev{search state}. At each replanning iteration \(k\), \(s_k\) \rev{summarizes solution quality}, feasibility, \rev{stagnation}, remaining budget, and \rev{tool feedback from the recorded trajectory}\rev{. This representation connects the observed search progress to the next optimization intention}. Based on this representation, the strategy planner produces
% \begin{equation}
% \small
% z_k
% =
% \Phi_{\mathrm{LLM}}(s_k)
% \label{eq:strategy_planning}
% \end{equation}
$
z_k
=
\Phi_{\mathrm{LLM}}(s_k)$,
\label{eq:strategy_planning}
where \(\Phi_{\mathrm{LLM}}\) denotes an inference-time decision process induced by a prompt, and \(z_k\) denotes the optimization strategy selected at iteration \(k\).

\subsection{Strategy-Guided Adaptive Tool Planner}

To translate the strategy selected by the agent into concrete solution updates, we introduce a strategy-guided adaptive tool planner that balances tool exploration and optimization gain. The strategy \(z_k\) represents the optimization intention at iteration \(k\), but does not directly specify which tool should be executed. 
\paragraph{Candidate Tool Library Construction.}The planner first constructs a candidate tool library that is compatible with both the problem structure and the current strategy\rev{, given by}
% \begin{equation}
% \small
% \mathcal{C}_k
% =
% \mathcal{T}(P)\cap\mathcal{T}(z_k),
% \label{eq:candidate_tool_set}
% \end{equation}
$
\mathcal{C}_k
=
\mathcal{T}(P)\cap\mathcal{T}(z_k)$,
where \(\mathcal{T}(P)\) denotes the set of tools applicable to problem \(P\), and \(\mathcal{T}(z_k)\) denotes the set of tools consistent with strategy \(z_k\). Their intersection maps the abstract optimization strategy to a library of executable candidate tools.

% \paragraph{Balancing Tool Exploration and Optimization Gain.}
% % After constructing the candidate tool library, the planner balances the exploration of insufficiently evaluated tools against the exploitation of tools that have demonstrated high optimization gains.
% After constructing the candidate tool library, the planner balances exploration and exploitation by evaluating less-tested tools while prioritizing those with proven effectiveness.
% Let
% \(\mathcal{C}_k^{0}=\left\{t\in\mathcal{C}_k \mid n_t=0\right\}\)
% denote the set of untried candidate tools, and let \(\mathcal{C}_k^{\min}\) denote the set of candidate tools with the smallest invocation count, where \(n_t\) is the number of times tool \(t\) has been invoked. At iteration \(k\), the tool to be executed is planned according to
% \begin{equation}
% t_k
% =
% \begin{cases}
% \operatorname{Sample}\!\left(\mathcal{C}_k^{0}\right),
% & \mathcal{C}_k^{0}\neq\varnothing,\\[4pt]
% \operatorname{Sample}\!\left(\mathcal{C}_k^{\min}\right),
% & \text{with probability }\epsilon,\\[4pt]
% \displaystyle
% \arg\max_{t\in\mathcal{C}_k}p_k(t),
% & \text{otherwise}.
% \end{cases}
% \label{eq:tool_selection}
% \end{equation}

% Equation~\eqref{eq:tool_selection} balances tool exploration and gain-oriented exploitation through three complementary branches. If untried tools remain, the planner samples from \(\mathcal{C}_k^{0}\), ensuring that every strategy-compatible tool is evaluated at least once. After all candidate tools have been invoked, it continues to explore with probability \(\epsilon\) by sampling from \(\mathcal{C}_k^{\min}\). 

\paragraph{Balancing Tool Exploration and Optimization Gain.}
After constructing the candidate tool library, the planner balances
exploration and exploitation by evaluating less-tested tools while
prioritizing those with proven effectiveness. Let
\(
\mathcal{C}_k^{0}
=
\left\{t\in\mathcal{C}_k \mid n_t=0\right\}
\)
denote the set of untried candidate tools, and define
\(
\mathcal{C}_k^{\min}
=
\left\{
t\in\mathcal{C}_k
\;\middle|\;
n_t=\min_{s\in\mathcal{C}_k} n_s
\right\}
\)
as the set of candidate tools with the smallest invocation count.
Here, \(n_t\) denotes the number of times tool \(t\) has been invoked.
We further define the gain-maximizing tool as \(t_k^{\star}=\arg\max_{t\in\mathcal{C}_k} p_k(t)\), where \(p_k(t)\) denotes the priority score assigned to tool \(t\) at
iteration \(k\).

Conditioned on the optimization trajectory \(s_k\), the tool
\(t_k\) is selected according to the following probability distribution
\begin{equation}
\small
\Pr\!\left(t_k=t \mid s_k\right)
=
\begin{cases}
\displaystyle
\frac{\mathbb{I}\!\left[t\in\mathcal{C}_k^{0}\right]}
     {\left|\mathcal{C}_k^{0}\right|},
& \mathcal{C}_k^{0}\neq\varnothing,\\[10pt]
\begin{aligned}
&\displaystyle
\epsilon
\frac{\mathbb{I}\!\left[t\in\mathcal{C}_k^{\min}\right]}
     {\left|\mathcal{C}_k^{\min}\right|}\\[-1pt]
&\displaystyle\quad
+(1-\epsilon)
\mathbb{I}\!\left[t=t_k^{\star}\right]
\end{aligned}
& \mathcal{C}_k^{0}=\varnothing.
\end{cases}
\label{eq:tool_selection}
\end{equation}
where \(\mathbb{I}[\cdot]\) is the indicator function and
\(\epsilon\in[0,1]\) controls the exploration rate. 
If untried tools remain, the planner selects one uniformly from \(\mathcal{C}_k^{0}\), prioritizing the evaluation of previously unused candidates. Once all current candidate tools have been invoked, the planner explores with probability \(\epsilon\) by selecting uniformly from the least-invoked tools in \(\mathcal{C}_k^{\min}\). With probability \(1-\epsilon\), the planner follows the current optimization trajectory by selecting highest-priority tool \(t_k^{\star}\).
This mechanism maintains tool diversity and prevents premature concentration on a small subset of tools.

\paragraph{Dynamic Tool Priority \(p_k(t)\).}When randomized exploration is not performed, the planner selects the tool with the highest dynamic priority\rev{, defined as}
\begin{equation}
\small
p_k(t)
=
\alpha \bar{r}_t
+
(1-\alpha)r_t^{\max}
-
\lambda\zeta_t,
\label{eq:tool_priority}
\end{equation}
where \(\bar{r}_t\) and \(r_t^{\max}\) respectively denote the average and maximum historical gains of tool \(t\), and \(\alpha\in[0,1]\) controls their relative contributions. Moreover, \(\zeta_t\) is the number of consecutive invocations without improvement, and \(\lambda\geq0\) controls the stagnation penalty. Once the tool produces an improvement, \(\zeta_t\) is reset to zero.

The planner thus maintains tool diversity through initial and randomized exploration while progressively allocating more resources to tools that produce larger and more reliable improvements.

\subsection{Probabilistic Local Convergence Analysis}

At each replanning time, the agent maps the optimization trajectory to a strategy $z$. Rather than updating the candidate solutions directly, TASA activates the compatible tool library $\mathcal{C}(z):=\mathcal{T}(P)\cap\mathcal{T}(z)$, from which the tool planner selects an executable tool. We first characterize the neighborhoods induced by the strategies
and their compatible tools, and then establish the corresponding
local convergence result.

\paragraph{Definition 1 (Strategy-Conditioned Tool Neighborhood)}
For each $t \in \mathcal{C}(z)$, let $q_t(x' \mid x)$ denote the probability that tool $t$ generates a feasible candidate $x'$ using the reference solution $x$. The neighborhood induced by strategy $z$ is
\begin{equation}
\small
    \mathcal{N}_{\mathcal{T}}(x\mid z)
    :=\bigcup_{t\in\mathcal{C}(z)}
      \{x'\in\mathcal{X}_P:q_t(x'\mid x)>0\}.
\end{equation}

Since TASA may activate different strategies, we aggregate their
neighborhoods to define the overall local search region.

\paragraph{Definition 2 (TASA-Induced Local Optimum)}
Define the overall reachable neighborhood $\mathcal{N}_{\mathrm{TASA}}(x)$ and the improving set $\mathcal{I}_{\mathrm{TASA}}(x)$ of TASA, respectively, as
\begin{align}
\small
    \mathcal{N}_{\mathrm{TASA}}(x)
    &:=\bigcup_{z\in\mathcal{Z}}\mathcal{N}_{\mathcal{T}}(x\mid z),\\
    \mathcal{I}_{\mathrm{TASA}}(x)
    &:=\{x'\in\mathcal{N}_{\mathrm{TASA}}(x):
      f_P(x')>f_P(x)\}.
\end{align}
A feasible solution is TASA-locally optimal if
$\mathcal{I}_{\mathrm{TASA}}(x)=\varnothing$. The set of all TASA-locally
optimal solutions is denoted by $\mathcal{L}_{\mathrm{TASA}}$.

At iteration $k$, $\hat{x}_P^{k}$ denotes the best feasible solution found so far. $\mathcal{B}_k$ denotes the generated batch. $x_k$ denote the reference solution. $x'_{k}$ denote the candidate solution.

\paragraph{Assumption 3 (Strict Improvements)}
The feasible set $\mathcal{X}_P$ is finite. \rev{Assume an initial feasible solution is available}\rev{.} TASA updates
\begin{equation}
\small
    \hat{x}_P^{k+1}
    =
    \arg\max_{x\in\{\hat{x}_P^{k}\}\cup(\mathcal{B}_k\cap\mathcal{X}_P)}
    f_P(x),
    \label{eq:elite-update}
\end{equation}
breaking ties in favor of $\hat{x}_P^k$. Thus, the best feasible solution changes only after a strict improvement. Since strict improvements can occur only finitely many times, persistent exploration is needed to ensure that TASA does not remain at a solution that is not locally optimal.

\paragraph{Assumption 4 (Persistent Local Exploration)}
\rev{Let $\mathcal{F}_k$ contain the full search history before batch $\mathcal{B}_k$}\rev{.} For \rev{every $x\notin\mathcal{L}_{\mathrm{TASA}}$}, there exists $p_x>0$\rev{, uniform over iterations and histories with $\hat{x}_P^k=x$}\rev{,} such that \rev{almost surely
\begin{equation}
\small
    \Pr\!\left(
      \mathcal{B}_k\cap\mathcal{I}_{\mathrm{TASA}}(x)\neq\varnothing
      \mid \mathcal{F}_k
    \right)\geq p_x.
    \label{eq:persistent}
\end{equation}}

Assumptions 3 and 4 ensure finite improvement steps and exclude non-locally optimal solutions as limiting states. To formalize these implications, we establish two auxiliary lemmas, whose statements and proofs are deferred to supplementary material.

% \paragraph{Lemma 1 (Finite Strict Improvements)}
% Under Assumption 1, TASA accepts at most $|f_P(\mathcal{X}_P)|-1$ strict improvements.

% \paragraph{Lemma 2 (Non-Locally Optimal Solutions Are Not Absorbing)}
% Under Assumption 2, if $x_\tau=x\notin\mathcal{L}_{\mathrm{TASA}}$ at a stopping time $\tau$, then
% the conditional probability of remaining at $x$ forever is zero.

\paragraph{\rev{Theorem} 5 (TASA-Local Convergence)}
Under Assumptions 3 and 4, the probability that the best feasible
solution reaches a TASA-induced local optimum in finite time and
remains unchanged thereafter is given by
\begin{equation}
\small
\begin{aligned}
\Pr\!\bigl(
    &\exists\, K<\infty,\;
    \exists\, x_\infty\in\mathcal{L}_{\mathrm{TASA}},\\
    &\text{such that }\hat{x}_P^k=x_\infty
    \text{ for all } k\ge K
\bigr)=1.
\end{aligned}
\end{equation}
Here, $K$ is a finite random time and $x_\infty$ is the random
limiting best feasible solution. The proof \rev{appears} in the supplementary material. \rev{Assumption 4 requires} persistent exploration \rev{through strategy planning} and \rev{tool execution}\rev{, whereas \eqref{eq:tool_selection} describes tool selection for the current strategy}\rev{. Together with} strict improvement\rev{, this assumption yields local} convergence.

\begin{table*}[t]
\vspace{-30pt}
\centering
\setlength{\tabcolsep}{3pt}
\renewcommand{\arraystretch}{1.12}
\small

% =========================================================
% First four benchmark tasks
% =========================================================
\resizebox{\textwidth}{!}{%
\begin{tabular}{c|l|c|c|c|c}
\toprule

\textbf{Category}
&
\textbf{Method}
&
\textbf{KP}
&
\textbf{MaxCut}
&
\textbf{BGP}
&
\textbf{TSP}
\\

\midrule

\shortstack{
\textbf{Approximate}
\\
\textbf{Global-Optimum}
}
&
\raisebox{0.8\height}{RefSolver}
&
\raisebox{0.8\height}{$\mathbf{1.227 \pm 0.069}$}
&
\raisebox{0.8\height}{$1.001 \pm 0.163$}
&
\raisebox{0.8\height}{$\mathbf{1.912 \pm 0.118}$}
&
\raisebox{0.8\height}{$\underline{1.693 \pm 0.097}$}
\\

\midrule

\multirow{4}{*}{
\shortstack{
\textbf{Traditional}\\
\textbf{Optimization}
}}
&
RandSearch
&
$-1.510 \pm 0.255$
&
$-0.650 \pm 0.302$
&
$-0.635 \pm 0.186$
&
$-0.754 \pm 0.048$
\\

&
BO
&
$-0.053 \pm 0.086$
&
$-0.501 \pm 0.255$
&
$-0.353 \pm 0.077$
&
$-0.719 \pm 0.054$
\\

&
GA
&
$-0.855 \pm 0.342$
&
$-0.469 \pm 0.196$
&
$-0.122 \pm 0.152$
&
$-0.687 \pm 0.060$
\\

&
CB
&
$\underline{1.221 \pm 0.063}$
&
$-0.617 \pm 0.204$
&
$-0.635 \pm 0.186$
&
$1.365 \pm 0.064$
\\

\midrule

\multirow{4}{*}{
\shortstack{
\textbf{Direct LLM}\\
\textbf{Solver}
}}
&
OPRO
&
$0.008 \pm 0.286$
&
$-1.033 \pm 0.173$
&
$-1.069 \pm 0.136$
&
$-0.709 \pm 0.082$
\\

&
E2E Solver
&
$0.235 \pm 0.179$
&
\textemdash
&
\textemdash
&
$-0.870 \pm 0.014$
\\

&
Block E2E Solver
&
$0.094 \pm 0.119$
&
$1.148 \pm 0.117$
&
$-0.109 \pm 0.169$
&
$0.364 \pm 0.016$
\\

&
Block-LS
&
$0.463 \pm 0.022$
&
$\underline{1.183 \pm 0.176}$
&
$-0.100 \pm 0.181$
&
$0.368 \pm 0.053$
\\

\midrule

\multirow{3}{*}{
\shortstack{
\textbf{LLM-Generated}\\
\textbf{Heuristic}
}}
&
FunSearch
&
$-0.526 \pm 0.414$
&
$-0.309 \pm 0.336$
&
$0.009 \pm 0.248$
&
$-0.704 \pm 0.047$
\\

&
EoH
&
$-0.023 \pm 1.845$
&
$0.297 \pm 1.081$
&
$0.676 \pm 0.982$
&
$-0.744 \pm 0.104$
\\

&
CoT
&
$-1.510 \pm 0.255$
&
$-1.509 \pm 0.638$
&
$-1.083 \pm 0.203$
&
$-0.551 \pm 0.053$
\\

\midrule

\textbf{Ours}
&
\textbf{TASA}
&
$\mathbf{1.227 \pm 0.069}$
&
$\mathbf{1.459 \pm 0.009}$
&
$\underline{1.903 \pm 0.132}$
&
$\mathbf{1.949 \pm 0.051}$
\\

\bottomrule
\end{tabular}%
}

\vspace{3mm}

% =========================================================
% Last four benchmark tasks
% =========================================================
\resizebox{\textwidth}{!}{%
\begin{tabular}{c|l|c|c|c|c}
\toprule

\textbf{Category}
&
\textbf{Method}
&
\textbf{VRP}
&
\textbf{PDR}
&
\textbf{JSSP}
&
\textbf{MS}
\\

\midrule

\shortstack{
\textbf{Approximate}
\\
\textbf{Global-Optimum}
}
&
\raisebox{0.8\height}{RefSolver}
&
\raisebox{0.8\height}{$\underline{2.290 \pm 0.069}$}
&
\raisebox{0.8\height}{$\mathbf{2.740 \pm 0.063}$}
&
\raisebox{0.8\height}{$\underline{1.690 \pm 0.372}$}
&
\raisebox{0.8\height}{$\underline{1.920 \pm 0.361}$}
\\

\midrule

\multirow{4}{*}{
\shortstack{
\textbf{Traditional}\\
\textbf{Optimization}
}}
&
RandSearch
&
$-0.452 \pm 0.013$
&
$-0.718 \pm 0.087$
&
$-0.250 \pm 0.111$
&
$-0.009 \pm 0.271$
\\

&
BO
&
$-0.401 \pm 0.038$
&
$-0.288 \pm 0.174$
&
$-0.248 \pm 0.138$
&
$-0.241 \pm 0.514$
\\

&
GA
&
$-0.439 \pm 0.013$
&
$-0.209 \pm 0.171$
&
$-0.428 \pm 0.206$
&
$-0.272 \pm 0.165$
\\

&
CB
&
$-0.452 \pm 0.013$
&
$-0.718 \pm 0.087$
&
$-1.287 \pm 0.160$
&
$-0.009 \pm 0.271$
\\

\midrule

\multirow{4}{*}{
\shortstack{
\textbf{Direct LLM}\\
\textbf{Solver}
}}
&
OPRO
&
$-0.405 \pm 0.016$
&
$-0.039 \pm 0.249$
&
$-0.372 \pm 0.310$
&
$-0.939 \pm 0.399$
\\

&
E2E Solver
&
$-0.487 \pm 0.021$
&
$-0.464 \pm 0.613$
&
$-1.356 \pm 0.601$
&
$-0.939 \pm 0.399$
\\

&
Block E2E Solver
&
$-0.441 \pm 0.028$
&
$-0.138 \pm 0.246$
&
$-0.402 \pm 0.209$
&
$-0.939 \pm 0.399$
\\

&
Block-LS
&
$-0.413 \pm 0.026$
&
$-0.047 \pm 0.135$
&
$-0.347 \pm 0.182$
&
$-0.772 \pm 0.114$
\\

\midrule

\multirow{3}{*}{
\shortstack{
\textbf{LLM-Generated}\\
\textbf{Heuristic}
}}
&
FunSearch
&
$-0.375 \pm 0.022$
&
$-0.516 \pm 0.152$
&
$-0.269 \pm 0.211$
&
$-0.412 \pm 0.518$
\\

&
EoH
&
$-0.325 \pm 0.039$
&
$-0.480 \pm 0.413$
&
$0.449 \pm 0.815$
&
$0.601 \pm 0.713$
\\

&
CoT
&
$-0.492 \pm 0.009$
&
$-0.718 \pm 0.087$
&
$0.917 \pm 0.391$
&
$-0.009 \pm 0.271$
\\

\midrule

\textbf{Ours}
&
\textbf{TASA}
&
$\mathbf{2.393 \pm 0.068}$
&
$\underline{1.595 \pm 0.138}$
&
$\mathbf{1.903 \pm 0.383}$
&
$\mathbf{2.020 \pm 0.285}$
\\

\bottomrule
\end{tabular}%
}

\caption{Performance comparison on eight benchmark tasks.
Results are reported as mean $z$-score $\pm$ standard deviation over
seeds 1--3. Higher values indicate better performance. The best result
for each task is shown in \textbf{bold}, and the second-best result is
\underline{underlined}. \textemdash\ indicates that no feasible
solution was obtained for at least one seed.}
\vspace{-15pt}
\label{tab:hard_benchmark_results}
\end{table*}

\section{Experiments}
\subsection{Experimental Setup}
\paragraph{Datasets.}
To comprehensively evaluate our method, we consider eight representative combinatorial optimization problems in two groups. The first group includes discrete structure optimization problems\rev{, namely} the 0--1 Knapsack Problem (KP), Maximum Cut Problem (MaxCut), Balanced Graph Partition Problem (BGP), and Traveling Salesman Problem (TSP). These problems assess the adaptability of our method to diverse solution representations and combinatorial search spaces. The second group covers routing and scheduling problems\rev{, namely} the Capacitated Vehicle Routing Problem (VRP), Pickup-Delivery Routing Problem (PDR), Job-Shop Scheduling Problem (JSSP), and Unrelated Parallel-Machine Scheduling Problem (MS). They evaluate the method's ability to handle complex constraints, including capacity limits, precedence relations, resource competition, and load balancing. Detailed instance sizes, parameter settings, and generation procedures are provided in the supplementary material.

\paragraph{Baselines.}
We compare our method with twelve baselines, including one approximate global-optimum baseline, four traditional optimization baseline, four direct LLM baseline, and three LLM-generated heuristic baseline.

\noindent\textbf{Approximate Global-Optimum Baseline.}
\textbf{RefSolver} (\emph{Reference Solver})~\cite{perron2023refsolver} applies an exact algorithm when tractable and otherwise returns a solution produced by a deterministic strong heuristic.
\noindent\textbf{Traditional optimization baselines.}
\textbf{RandSearch} (\emph{Random Search})~\cite{jame2012randsearch} independently samples candidate solutions without exploiting evaluation history.
\textbf{BO} (\emph{Bayesian Optimization}) uses a lightweight kernel-regression model to estimate objective values and selects candidates based on an upper confidence bound.
\textbf{GA} (\emph{Genetic Algorithm}) maintains a population of candidate solutions and generates new candidates through selection, crossover, and mutation.
\textbf{CB} (\emph{Combinatorial Bandit}) treats solution components as bandit arms and constructs candidates according to their estimated rewards and exploration bonuses.

\noindent\textbf{Direct LLM solver baselines.}
\textbf{OPRO} (\emph{Optimization by PROmpting})~\cite{yang2024OPRO} prompts the LLM with the problem description and previously evaluated solutions to directly generate complete candidate solutions.
\textbf{E2E Solver} (\emph{End-to-End LLM Solver})~\cite{jiang2025E2ESolver} uses a fixed prompt to directly generate complete candidate solutions for problem instance, without using previously evaluated solutions or optimization feedback.
\textbf{Block E2E Solver} (\emph{Block-wise End-to-End LLM Solver})~\cite{zheng2024BlockE2E} decomposes a high-dimensional solution into blocks and asks the LLM to update one block at a time.
\textbf{Block-LS} (\emph{Block-wise LLM Solver with Local Search})~\cite{jiang2025E2ESolver,zheng2024BlockE2E} alternates block-wise LLM candidate generation with local refinement.

\noindent\textbf{LLM-generated heuristic baselines.}
\textbf{FunSearch}~\cite{romera2024FunSearch} is a lightweight archive-guided search proxy that periodically injects randomly generated candidates and otherwise mutates high-scoring historical solutions.
\textbf{EoH} (\emph{Evolution of Heuristics})~\cite{liu2024EoH} periodically introduces random candidates and otherwise recombines and mutates high-scoring historical solutions.
\textbf{CoT} (\emph{CoT Direct Solver})~\cite{sprague2025cot} uses chain-of-thought prompting to guide the LLM through step-by-step problem analysis and directly generate candidate solutions. Detailed implementation settings are provided in the supplementary material.
\rev{The supplementary material details these implementations}\rev{, including LLM-call limits and subsequent local-search operations}\rev{.}

\paragraph{Implementation Details.}
Qwen-Plus~\cite{yang2025qwen3} is accessed via API and used as the language model. Each experiment is conducted using random seeds 1, 2, and 3. Under each seed, all methods are evaluated on the same problem instance, with randomness in instance generation and initialization controlled by the corresponding seed. Each run is limited to 60 candidate solution evaluations with a batch size of 2\rev{. 
The budget counts generated candidates}\rev{. Tool construction}\rev{, LLM calls}\rev{, and computations within tools depend on the method}\rev{, so results compare solution quality under this budget and the reported implementations}.

\paragraph{Metrics.}
For fair comparison across problems with different objective scales and optimization directions, all native objectives are converted into higher-is-better scores. For problem \(p\), seed \(s\), and method \(m\), the best feasible score is standardized as
$
z_{p,s,m}
=
\frac{y_{p,s,m}-\mu_{p,s}}{\sigma_{p,s}}$.
Here, \(y_{p,s,m}\) denotes the best feasible value of \(f_P(x)\) achieved by method \(m\). Details of the computation of \(f_P(x)\) are provided in the supplementary material. For each problem \(p\) and seed \(s\), \(\mu_{p,s}\) is the mean of these values across all methods that produce a feasible solution, and \(\sigma_{p,s}\) is the corresponding population standard deviation. Accordingly, \(z_{p,s,m}>0\) indicates above-average performance, whereas \(z_{p,s,m}<0\) indicates below-average performance for the corresponding problem and seed.

\subsection{Comparative Performance on Benchmark Tasks}

From the results presented in Table~\ref{tab:hard_benchmark_results}, it can be observed that TASA achieves the strictly best mean score on five tasks, ties RefSolver for the best result on KP, and ranks second on BGP and PDR. On BGP, it nearly matches RefSolver, with a gaps of only 0.009. Its average \(z\)-score across all tasks is 1.806, almost identical to RefSolver's 1.809. These results show that, under the same evaluation budget, TASA is competitive with specialized solvers while retaining a unified framework across problems with diverse structures and constraints.

TASA also consistently outperforms direct LLM-based methods. It exceeds OPRO on every task by 2.376 \(z\)-score points on average and surpasses the strongest direct LLM baseline by an average of 1.762 points. The advantage is particularly clear on BGP, VRP, JSSP, and MS, where direct LLM solvers obtain negative mean scores while TASA remains strongly positive. These negative values arise from \(z\)-score normalization, as a method receives a negative score when its best feasible score falls below the cross-method mean for the corresponding problem and seed. These results \rev{support} the \rev{combined use} of optimization trajectories\rev{, strategy planning}\rev{,} and problem-specific tools \rev{in TASA under the reported settings}.

Finally, TASA finds feasible solutions for every seed on all eight tasks and has a standard deviation below 0.2 on six tasks, demonstrating reliable performance across repeated runs. Because some baselines do not reliably produce feasible solutions across random seeds, we use ``\textemdash{}'' in Table~\ref{tab:hard_benchmark_results} to indicate that no feasible solution was obtained for at least one seed. On four of the five tasks where it ranks first, its standard deviation is no larger than that of RefSolver. Its main weakness is PDR, where it trails RefSolver by 1.145 points, suggesting that more specialized routing and repair tools are needed. Overall, the results demonstrate the effectiveness, generality, and robustness of our method.

% 消融
% \begin{figure}[t]
%     \centering
%     \includegraphics[width=0.9\columnwidth]{pic2.png}
%     \caption{Traceability of TASA over 60 evaluations on a representative run of VRP (Seed 3). The curve shows the best-so-far scores, while the markers and colored timelines indicate the optimization strategies selected throughout the search.}
%     \label{fig:2}
% \end{figure}

% \begin{figure}[t]
%     \centering
%     \includegraphics[width=\columnwidth]{pic3.png}
%     \caption{Effect of the tool-planning mechanism on best-so-far performance over 60 evaluations for TSP and VRP. The curves show the mean best-so-far scores over seeds 1-3, and the shaded regions indicate one standard deviation around the mean. Ours denotes the strategy-guided adaptive tool planner, while Random and UCB1 are alternative tool-planning policies.}
%     \label{fig:3}
% \end{figure}
\begin{figure}[t]
\vspace{-30pt}
    \centering

    \begin{minipage}{0.48\columnwidth}
        \centering
        \includegraphics[width=\linewidth]{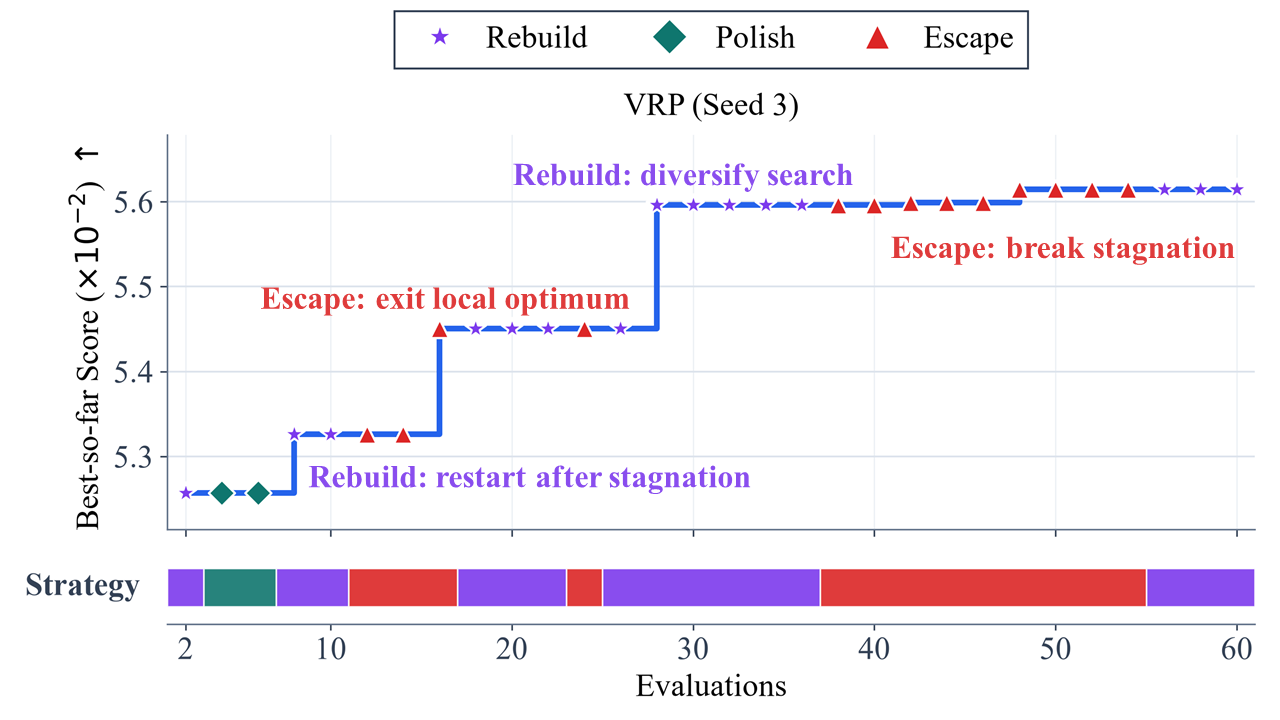}
        \captionof{figure}{Traceability of TASA over 60 evaluations on a representative run of VRP (Seed 3). The curve shows the best-so-far scores, while the markers and colored timelines indicate the optimization strategies selected throughout the search.}
        \label{fig:2}
    \end{minipage}
    \vspace{-5pt}
    \hfill
    \begin{minipage}{0.48\columnwidth}
        \centering
        \includegraphics[width=\linewidth]{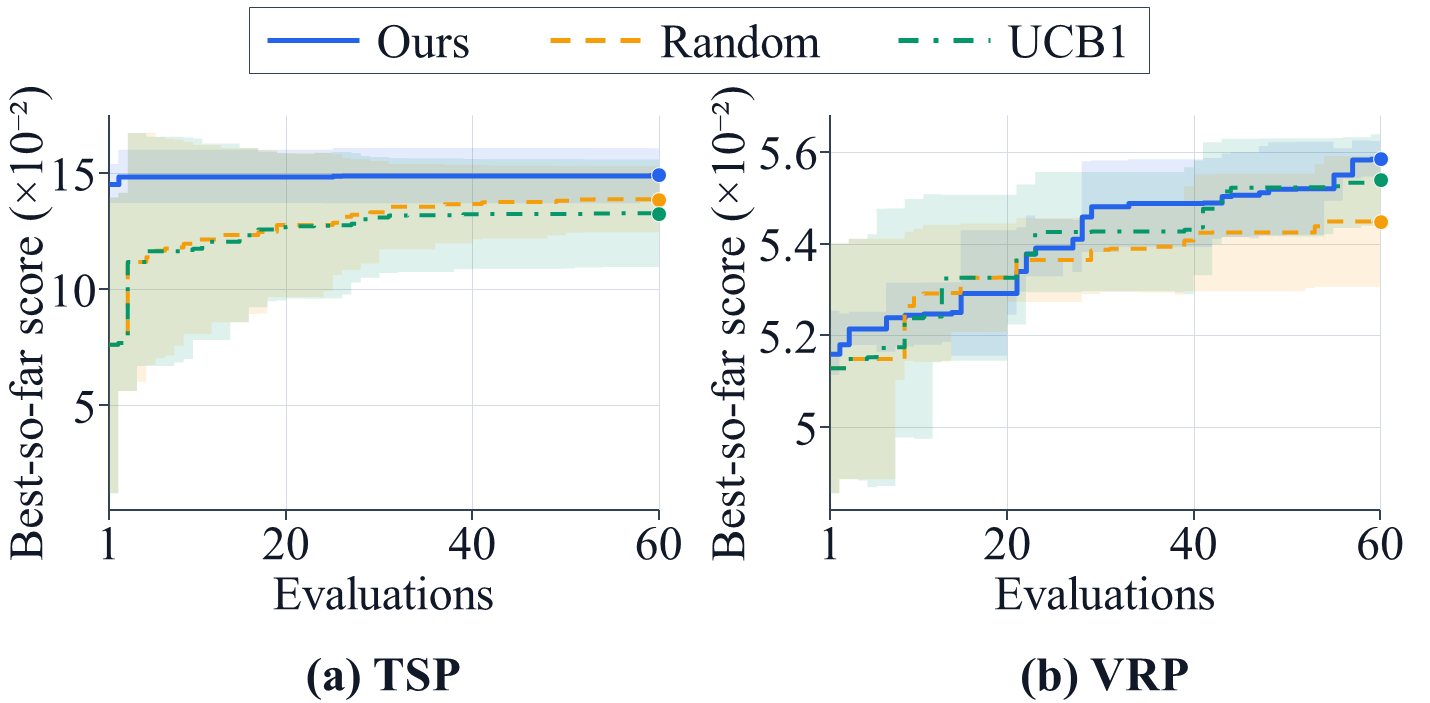}
        \captionof{figure}{Effect of the tool-planning mechanism on best-so-far performance over 60 evaluations for TSP and VRP. The curves show the mean best-so-far scores over seeds 1-3, and the shaded regions indicate one standard deviation around the mean. Ours denotes the strategy-guided adaptive tool planner, while Random and UCB1 are alternative tool-planning policies.}
        \label{fig:3}
    \end{minipage}
    \vspace{-5pt}

\end{figure}

% \begin{figure}[t]
%     \centering
%     \includegraphics[width=\columnwidth]{pic4.png}
%     \caption{Effect of tool-library size on best-so-far performance over 60 evaluations for JSSP and VRP. The curves show the mean scores over seeds 1--3, and the shaded regions indicate one standard deviation around the mean. All Tools uses the full library, while Half Tools and Quarter Tools retain one-half and one-quarter of the tools, respectively.}
%     \label{fig:4}
% \end{figure}

% \begin{figure}[t]
%     \centering
%     \includegraphics[width=\columnwidth]{pic5.png}
%     \caption{Effect of feedback granularity on best-so-far performance for BGP and MaxCut. The bars report the mean best-so-far scores over seeds 1–3 at evaluations 1, 20, 40, and 60. Full Feedback provides the strategy planner with the complete optimization trajectory, whereas Solution-Level Feedback retains only solution-level information.}
%     \label{fig:5}
% \end{figure}
\begin{figure}[t]
% Avoid covering the preceding figure captions with this float.
\vspace{0pt}
    \centering

    \begin{minipage}{0.48\columnwidth}
        \centering
        \includegraphics[width=\linewidth]{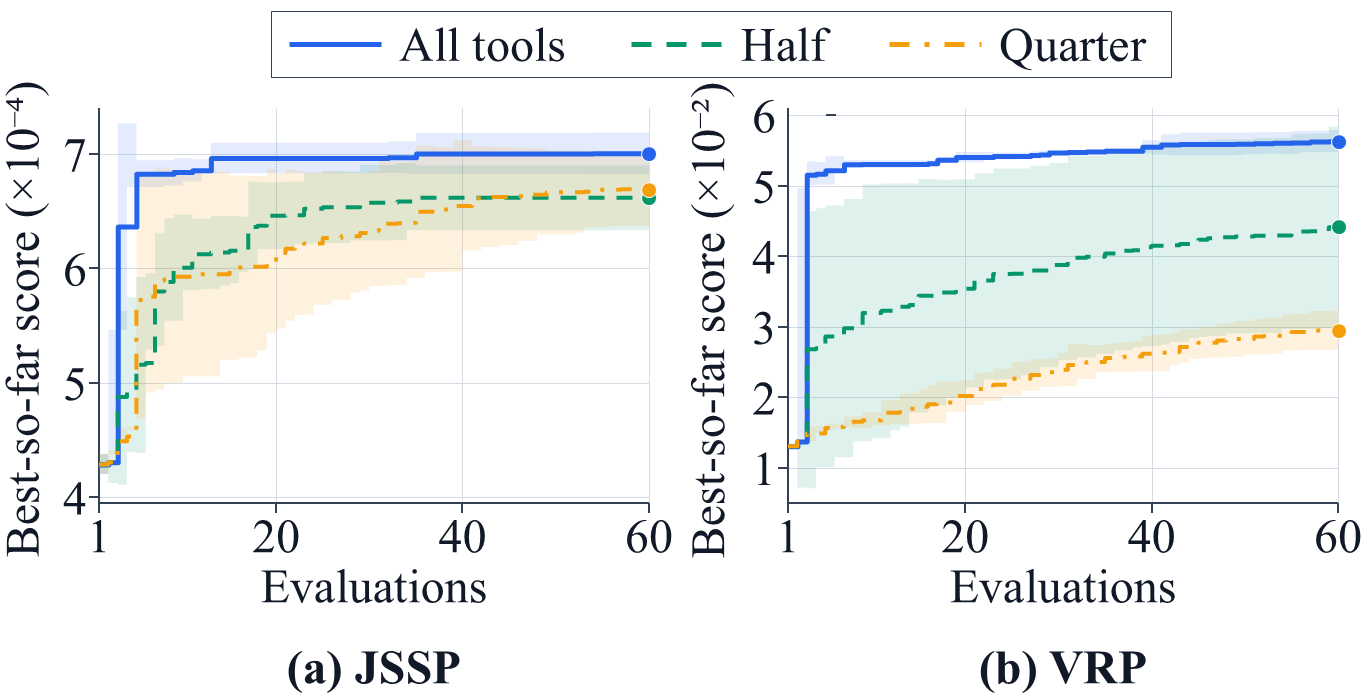}
        \captionof{figure}{Effect of tool-library size on best-so-far performance over 60 evaluations for JSSP and VRP. The curves show the mean scores over seeds 1--3, and the shaded regions indicate one standard deviation around the mean. All Tools uses the full library, while Half Tools and Quarter Tools retain one-half and one-quarter of the tools, respectively.}
        \label{fig:4}
    \end{minipage}
    \vspace{-5pt}
    \hfill
    \begin{minipage}{0.48\columnwidth}
        \centering
        \includegraphics[width=\linewidth]{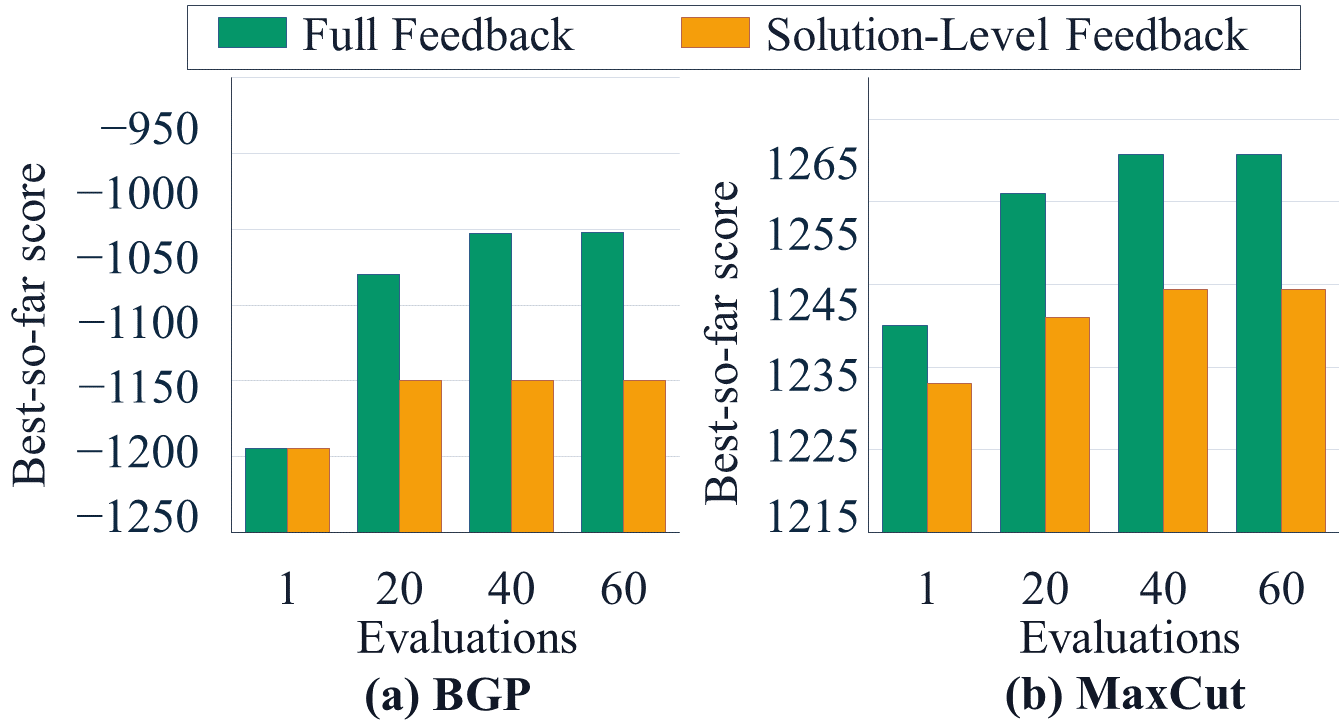}
        \captionof{figure}{Effect of feedback granularity on best-so-far performance for BGP and MaxCut. The bars report the mean best-so-far scores over seeds 1--3 at evaluations 1, 20, 40, and 60. Full Feedback provides the strategy planner with the \rev{recorded} optimization trajectory, whereas Solution-Level Feedback retains only solution-level information.}
        \label{fig:5}
    \end{minipage}
    \vspace{-5pt}

\end{figure}

% \subsection{Traceability of TASA.} To examine the traceability of TASA during optimization process, we visualize the strategy timelines of two runs on VRP and JSSP. On VRP. As shown in Figure \ref{fig:2} when the search stagnates, TASA shifts from \emph{polish} to \emph{rebuild} and then to \emph{escape}, yielding successive improvements. On JSSP, TASA first selects \emph{repair}, then shifts to \emph{polish} based on the optimization trajectory, and finally selects \emph{escape} after prolonged stagnation. These examples show that TASA's improvements can be traced to explicit strategy planning, and structured tool executions.
\paragraph{Traceability of TASA.}
\rev{Figure}\rev{~}\rev{2 makes} strategy \rev{selections and subsequent score improvements traceable in} a representative VRP run. \rev{Rebuild precedes improvements near evaluations 8 and 28}, and \rev{escape} near \rev{16} and \rev{49}.

\subsection{Effect of the Tool Planner.}
To evaluate the effectiveness of the tool-planning mechanism, we compare our strategy-guided adaptive tool planner with random selection and UCB1. As shown in Figure \ref{fig:3}, our planner achieves the highest final scores on both TSP and VRP. On TSP, it rapidly reaches 14.92, whereas Random and UCB1 converge to only 13.88 and 13.27. These results \rev{support using historical tool gains and consecutive invocations without improvement to select} effective tools \rev{for} the intended \rev{optimization} strategy.

\subsection{Effect of Tool Library Size.}
To evaluate the effect of tool-library size, we compare TASA using all tools with variants retaining one-half or one-quarter of the library. As shown in Figure \ref{fig:4}, the full library achieves the best final scores on both JSSP and VRP, reaching 7.01 and 5.63, respectively. The advantage is especially pronounced on VRP, where using half and one-quarter of the tools yields only 4.42 and 2.95. \rev{The} larger \rev{tool library provides more problem-specific} tools for \rev{executing the selected strategies}, \rev{supporting the contribution} of tool diversity to solution quality.

\subsection{Effect of Feedback Granularity.}
To evaluate the effect of feedback granularity, we compare Full Feedback with Solution-Level Feedback. As shown in Figure \ref{fig:5}, Full Feedback achieves higher scores on both BGP and MaxCut. On BGP, Solution-Level Feedback plateaus by 20th evaluation, whereas Full Feedback continues to improve in subsequent evaluations. On MaxCut, both variants improve until around 40th evaluation and then change little, but Full Feedback consistently maintains higher scores. These results \rev{support using the} optimization \rev{trajectory to} provide \rev{the LLM strategy planner with} information \rev{about} search progress and stagnation.

\section{Conclusion}

In this paper, we propose TASA, which shifts the role of the LLM from directly generating solutions to planning optimization strategies based on optimization trajectories and executing them through a strategy-guided adaptive tool planner. Experimental results on eight benchmark problems demonstrate that TASA provides an effective and traceable framework for LLM-based combinatorial optimization.

\bibliography{iclr2027_conference}
\bibliographystyle{iclr2027_conference}

\end{document}

%% file: tex_engine_compat.tex
\ifdefined\XeTeXversion
  
  \renewcommand{\sfdefault}{cmss}
  
  \usepackage[OT1]{fontenc}
  \usepackage{newunicodechar}
  \newunicodechar{’}{\textquoteright}
  \newunicodechar{–}{\textendash}
\fi

%% file: math_commands.tex
\usepackage{amsmath,amsfonts,bm}

\def\eqref#1{equation~\ref{#1}}
\def\1{\bm{1}}

\DeclareMathAlphabet{\mathsfit}{\encodingdefault}{\sfdefault}{m}{sl}
\SetMathAlphabet{\mathsfit}{bold}{\encodingdefault}{\sfdefault}{bx}{n}